\documentclass{article} 
\usepackage{nips11submit_e,times}

\usepackage{graphicx}
\usepackage{amssymb}
\usepackage{epstopdf}
\usepackage{multirow}
\usepackage{algorithm}
\usepackage{algorithmic}
\usepackage{verbatim}
\usepackage{amsmath}
\usepackage{color}
\usepackage{rotating}

\newfont{\bboard}{msbm10 scaled\magstephalf}
\def\real{\mbox{\bboard R}}

\def\boardE{\mbox{\bboard E}}

\newcommand{\argmax}{ \mathop{\mathrm{argmax}} }
\newcommand{\EV}{\boardE}

\newcommand{\detbar}[1]{\left|#1\right|}
\newcommand{\norm}[1]{\left\|#1\right\|}

\newcommand{\Normalv}[3]{{\cal N}\left(#1\middle|#2,#3\right)}
\newcommand{\tNormalv}[3]{{\cal N}(#1|#2,#3)}

\DeclareMathOperator{\vect}{vec}

\newcommand{\cov}{\mathrm{cov}}

\newcommand{\mbf}[1]{\mathbf{#1}}
\newcommand{\diag}{\mathrm{diag}}

\newcommand{\pdd}[2]{\frac{\partial #1}{\partial #2}}

\newcommand{\tpdd}[2]{\tfrac{\partial #1}{\partial #2}}

\newcommand{\refeqn}[1]{(\ref{#1})}

\newcommand{\comments}[1]{}

\newcommand{\ie}{i.e. }

\newcommand{\va}{\mbf{a}}

\newcommand{\ve}{\mbf{e}}
\newcommand{\vx}{\mbf{x}}

\newcommand{\vg}{\mbf{g}}
\newcommand{\vy}{\mbf{y}}

\newcommand{\vk}{\mbf{k}}
\newcommand{\vz}{\mbf{z}}
\newcommand{\vm}{\mbf{m}}
\newcommand{\vt}{\mbf{t}}
\newcommand{\vT}{\mbf{T}}

\newcommand{\vu}{\mbf{u}}

\newcommand{\vone}{\mbf{1}}
\newcommand{\vzero}{\mbf{0}}
\newcommand{\valpha}{\boldsymbol{\alpha}}
\newcommand{\vgamma}{\boldsymbol{\gamma}}
\newcommand{\vomega}{\boldsymbol{\omega}}
\newcommand{\vtheta}{\boldsymbol{\theta}}
\newcommand{\veta}{\boldsymbol{\eta}}

\newcommand{\vmu}{\boldsymbol{\mu}}

\newcommand{\vpi}{\boldsymbol{\pi}}

\newcommand{\mA}{\mbf{A}}
\newcommand{\mB}{\mbf{B}}

\newcommand{\mJ}{\mbf{J}}
\newcommand{\mX}{\mbf{X}}
\newcommand{\mW}{\mbf{W}}
\newcommand{\mV}{\mbf{V}}
\newcommand{\mK}{\mbf{K}}

\newcommand{\mP}{\mbf{P}}

\newcommand{\mI}{\mbf{I}}
\newcommand{\mM}{\mbf{M}}
\newcommand{\mU}{\mbf{U}}
\newcommand{\mL}{\mbf{L}}

\newcommand{\mSigma}{\boldsymbol{\Sigma}}

\newcommand{\mTheta}{\boldsymbol{\Theta}}
\newcommand{\mPhi}{\boldsymbol{\Phi}}
\newcommand{\mGamma}{\boldsymbol{\Gamma}}
\newcommand{\mOmega}{\boldsymbol{\Omega}}

\newcommand{\mEta}{\mbf{H}}

\newcommand{\tveta}{\tilde{\veta}}

\newcommand{\tvz}{\tilde{\vz}}
\newcommand{\tvt}{\tilde{\vt}}

\newcommand{\vgammabyi}{\vgamma_{:}}
\newcommand{\vgammabyj}{\vgamma\s{:}}
\newcommand{\mGammabyi}{\mGamma_{:}}
\newcommand{\mGammabyj}{\mGamma\s{:}}
\newcommand{\mOmegabyi}{\mOmega_{:}}
\newcommand{\mOmegabyj}{\mOmega\s{:}}
\newcommand{\vetabyi}{\veta_{:}}
\newcommand{\vetabyj}{\veta\s{:}}
\newcommand{\vybyi}{\vy_{:}}
\newcommand{\vybyj}{\vy\s{:}}
\newcommand{\mWbyi}{\mW_{:}}
\newcommand{\mWbyj}{\mW\s{:}}
\newcommand{\mUbyi}{\mU_{:}}
\newcommand{\mUbyj}{\mU\s{:}}

\newcommand{\vtbyj}{\vt\s{:}}
\newcommand{\hvt}{\hat{\vt}}

\newcommand{\hvtbyj}{\hvt\s{:}}

\newcommand{\vzbyj}{\vz\s{:}}

\newcommand{\tvzbyj}{\tvz\s{:}}
\newcommand{\hvz}{\hat{\vz}}

\newcommand{\hvzbyj}{\hvz\s{:}}
\newcommand{\mAbyi}{\mA_{:}}

\newcommand{\mLbyi}{\mL_{:}}
\newcommand{\mLbyj}{\mL\s{:}}
\newcommand{\mMbyi}{\mM_{:}}
\newcommand{\mMbyj}{\mM\s{:}}

\newcommand{\hvu}{\hat{\vu}}

\newcommand{\hvubyj}{\hvu\s{:}}
\newcommand{\tvetabyi}{\tveta_{:}}
\newcommand{\tvetabyj}{\tveta\s{:}}

\newcommand{\tvu}{\tilde{\vu}}
\newcommand{\tmU}{\tilde{\mU}}
\newcommand{\tmW}{\tilde{\mW}}
\newcommand{\tvubyi}{\tvu_{:}}
\newcommand{\tvubyj}{\tvu\s{:}}

\newcommand{\tmUbyj}{\tmU\s{:}}
\newcommand{\tmWbyi}{\tmW_{:}}
\newcommand{\tmWbyj}{\tmW\s{:}}

\newcommand{\tvtbyj}{\tvt\s{:}}

\newcommand{\hmU}{\hat{\mU}}
\newcommand{\hmW}{\hat{\mW}}

\newcommand{\hmUbyj}{\hmU\s{:}}

\newcommand{\hmWbyj}{\hmW\s{:}}

\newcommand{\hveta}{\hat{\veta}}
\newcommand{\hvetabyi}{\hveta_{:}}
\newcommand{\hvetabyj}{\hveta\s{:}}

\newenvironment{smallbmatrix}{\left[\begin{smallmatrix}}{\end{smallmatrix}\right]}

\newcommand{\s}[1]{^{(#1)}}

\newcommand{\TODO}[1]{\textcolor{red}{[TODO: #1]}}

\title{Multivariate Generalized Gaussian Process Models}

\author{
Antoni B.~Chan\\
Department of Computer Science \\
City University of Hong Kong \\
Tat Chee Avenue, Kowloon Tong, Hong Kong\\
\texttt{abchan@cityu.edu.hk} 
}

\nipsfinalcopy 

\begin{document}

\maketitle
\vspace{-0.2in}

\begin{abstract}
\vspace{-0.15in}
We propose a {\em family} of multivariate Gaussian process models for correlated outputs, based on assuming that the likelihood function takes the generic form of the multivariate exponential family distribution (EFD).
We denote this model as a multivariate generalized Gaussian process model,
and derive Taylor and Laplace algorithms for approximate inference on the generic model.
By instantiating the EFD with specific parameter functions, we obtain two novel GP models (and corresponding inference algorithms) for correlated outputs: 1) a Von-Mises GP for  angle regression; and 2) a Dirichlet GP for regressing on the multinomial simplex.
\vspace{-0.15in}
\end{abstract}

\section{Introduction}

Gaussian process (GP) models are a non-parametric Bayesian approach to regression and classification \cite{GPML}.
In this work, we consider GP models with {\em multivariate outputs}, i.e, the vector observations.
A simple multivariate model applies a separate GP to each dimension, thus assuming statistical independence between the output dimensions.  However, this is contrary to actual multivariate observations, which tend to have significant correlation and structure.
To improve on this, GP models with correlated output can be formed by passing the independent GP priors through a function, such as
multi-class classification with the softmax \cite{Williams1998GPC} or probit functions \cite{Girolami05VBM,Kim2006emep},
ordinal regression with a multi-probit function \cite{Chu2005GPOR},
and
a semiparametric latent factor model \cite{Teh05SPLFM} with linear mixing.
%
Another line of research learns the correlation between output dimensions by  modeling the off-diagonal of the kernel block-matrix , e.g. multi-task  \cite{Chai2009MGP,Bonilla2008MGPP} and dependent GPs \cite{Boyle2005DGP}.
Finally, \cite{Salzmann2010CGP}  shows that multiple independent GPs that share the same kernel implicitly satisfy linear output constraints, and that higher-order constraints can be modeled by transforming the output vector accordingly, although this results in multiple predictions that require additional side-information to disambiguate.



In this paper, we follow the first line of work, and propose a {\em family} of multivariate GP models, which is based on assuming that the likelihood function takes the {\em generic} form of the {\em multivariate} exponential family distribution (EFD).
We call this generic model a {\em multivariate generalized Gaussian process model} (M-GGPM).
We derive two approximate inference algorithms for the generic M-GGPM, using the Taylor and Laplace approximations.  By instantiating the EFD with specific parameter functions, we obtain new GP regression models (and corresponding inference algorithms) for correlated outputs, e.g. the multinomials or angle vectors.
The proposed M-GGPM is a {\em non-trivial} multivariate extension of the univariate generalized GP model \cite{Chan2011CVPR}, as the M-GGPM needs additional care in handling degenerate covariances caused by constraints on the output vectors.

The remainder of the paper is organized as follows.  In \S \ref{text:mggpm} we present the multivariate GGPM, and in \S \ref{text:approxinf} we derive approximate inference for the general model.  Finally, in \S \ref{text:examples}, we present several examples of the model along with regression experiments.

\section{Multivariate Generalized Gaussian Process Models}
\label{text:mggpm}

In this section, we propose a general framework for multivariate GP models, based on the multivariate exponential family distribution.

\subsection{Multivariate Exponential Family Distribution}

The $d$-dimensional multivariate exponential family distribution (EFD) has the form
	\begin{align}
	\label{eqn:mexpo}
	p(\vy|\vtheta, \phi) = h(\vy, \phi) \exp \tfrac{1}{a(\phi)}\left(\vT(\vy)^T\vtheta - b(\vtheta) \right)
	\end{align}
where $\vtheta\in \real^D$ is a $D$-dimensional parameter vector, and  $\vy\in{\cal Y}$ is a $d$-dimensional observation vector from the set of possible observations ${\cal Y}$ (e.g., real vectors, multinomials, etc).  The function $\vT(\vy) : \real^d \rightarrow \real^D$ is a mapping of $\vy$ to the parameter space (i.e., sufficient statistic), and $\phi$ is a scalar dispersion parameter.
The function $b(\vtheta) : \real^D \rightarrow \real$ is the log-partition function, where
	$b(\vtheta) = \log  \int h(\vy,\phi) \exp(\vT(\vy)^T\vtheta) d\vy$.
The mean and covariance of $\vT(\vy)$ can be computed directly from the log-partition function as
	\begin{align}
	\label{eqn:mexpo_mean}
	\EV[\vT(\vy)|\vtheta] 
	= \nabla b(\vtheta)
	,
	\quad
	\cov(\vT(\vy)|\vtheta)
	= a(\phi)\nabla^2 b(\vtheta) ,
	\end{align}
where  $\nabla b(\vtheta) = \pdd{}{\vtheta}b(\vtheta)$ is the gradient, and $\nabla^2 b(\vtheta) = \pdd{}{\vtheta}\pdd{}{\vtheta^T} b(\vtheta)$ is the Hessian.
The multivariate EFD encompasses a wide variety of  distributions, such as the multivariate Gaussian, multinomial, Dirichlet, Von-Mises, and Wishart.
A specific distribution is characterized by the parameter functions $\mTheta = \{h(\vy,\phi), \vT(\vy), b(\vtheta), a(\phi)\}$.
Hence, our goal is to derive a multivariate GP model where the observation likelihood takes the {\em generic form} of the multivariate EFD, as parameterized by $\mTheta$.
After deriving the generic model, along with approximate inference algorithms, creating new types of multivariate GP models requires only plugging in a particular set of parameter functions $\mTheta$.

\subsection{Multivariate Generalized Gaussian Process Models (M-GGPM)}

We now consider a multivariate extension of  generalized GP models \cite{Chan2011CVPR}.
The model consists of three components: 1) a latent function with GP prior; 2) a noisy output, modeled as an EFD; and 3) a link function that relates the latent space and the output mean.  Formally, the M-GGPM is given by
	\begin{align}
	\veta(\vx) \sim {\cal GP}(\vzero, \mK(\vx,\vx')),
	\quad
	\vy \sim p(\vy|\vtheta, \phi),
	\quad
	\vg(\EV[\vT(\vy)|\vtheta]) = \veta(\vx).
	\end{align}
The latent function $\veta(\vx)$ is a $D$-dimensional vector-valued function, where $\eta\s{j}(\vx)$ is the jth dimension of the latent function (denoted by the super-script).  The latent function is modeled with a multivariate GP prior,
	$\veta(\vx) \sim {\cal GP}\left(\vzero, \mK(\vx,\vx') \right)$,
where $\mK(\vx,\vx')$ is the covariance {\em matrix} function.
In this paper, we will assume that the individual latent dimensions are uncorrelated, i.e., that the GP covariance matrix is diagonal,
	\begin{align}
	\mK(\vx,\vx') = \diag( k\s{1}(\vx,\vx'), \cdots, k\s{D}(\vx,\vx') )
	\end{align}
where $k\s{j}(\vx,\vx')$ is the covariance function for the jth latent dimension.
In this case, the dimensions of the latent space act as an ``independent factors'', which are later  combined by the EFD.
%

A noisy observation $\vy$ is distributed according to the multivariate EFD in \refeqn{eqn:mexpo}, i.e. $\vy \sim p(\vy|\vtheta, \phi)$, where $\vtheta$ is the parameter vector and $\phi$ is a hyperparameter.
Finally, the link function $\vg: \real^D \rightarrow \real^D$ maps the sufficient statistics of the observations, $\EV[\vT(\vy)|\vtheta]$, to the latent function $\veta(\vx)$.
Using \refeqn{eqn:mexpo_mean}, $\vtheta$ can be rewritten as a function of $\veta(\vx)$, yielding an equivalent formulation of the M-GGPM,
	\begin{align}
	\veta(\vx) \sim {\cal GP}(\vzero, \mK(\vx,\vx')),
	\quad
	\vy \sim p(\vy|\vtheta(\veta(\vx)), \phi),
	\quad
	\vtheta(\veta(\vx)) = [\nabla b]^{-1}\left(\vg^{-1}(\veta(\vx))\right).
	\end{align}
The link function that causes $\vtheta(\veta)=\veta$ is called the {\em canonical link function},
i.e. when $\vg(\vmu) = [\nabla b]^{-1}(\vmu)$ or $\vg(\veta)^{-1} = \nabla b(\veta)$.
The advantage with the link function is that it allows {\em direct specification} of the trend between the latent vector and the output mean.  While many trends can be represented by the kernel function (e.g., polynomial), there are some trends (e.g., logarithmic) that have no corresponding kernel function.  Hence, the link function is necessary in these cases.

\subsection{Notation}

$\{\vx_i,\vy_i\}$ is the ith training input/output pair,
and $\mX = [\vx_1 \cdots \vx_n]$ is the set of $n$ training inputs.
$\veta_i \in \real^D$ is the latent vector associated with data point $\vx_i$,
and $\mEta = [\veta_1 \cdots \veta_n] \in \real^{D \times n}$ is the matrix of latent vectors.
$\vetabyi = \vect(\mEta) \in \real^{nD}$ is the concatenation of all latent values ordered by data point.
$\veta\s{j} = \mEta^T \ve_j \in \real^n$ (the transpose of the jth row of $\mEta$) is the vector of all latent values of the jth dimension,
and $\vetabyj = \vect(\mEta^T)$ is the concatenation of all latent values ordered by dimension.
$\vetabyi$ and $\vetabyj$ are related through a commutation matrix $\mP$,
such that
$\vetabyj = \mP^T \vetabyi$ and $\vetabyi = \mP\vetabyj$.
In general, subscripts (e.g. $\veta_i$, $\vetabyi$) denote vectors or matrices with entries ordered by data point, while superscripts (e.g., $\veta\s{j}$, $\vetabyj$) denote ordering by latent dimension.

\subsection{Inference}

Given a training set $\{\vx_i,\vy_i\}_{i=1}^n$,
 the goal is to calculate a predictive  distribution of output $\vy_*$ given a novel input $\vx_*$.
The joint distribution of the latent vectors factors according to latent dimension,
	\begin{align}
	p(\vetabyj|\mX)
	&= \prod_{j=1}^D p(\veta\s{j}|\mX)
	= \prod_{j=1}^D \Normalv{\veta\s{j}}{\vzero}{\mK\s{j}},
	\end{align}
where $\mK\s{j}$ is the kernel matrix for the jth dimension, with corresponding kernel function $k\s{j}(\vx,\vx')$.
\comments{
the joint distribution of the latent variables is
	\begin{align}
	p\left(\begin{smallbmatrix}\vetabyj \\ \veta_*\end{smallbmatrix} | \mX, \vx_*\right)
	=
	\Normalv{\begin{smallbmatrix}\vetabyj \\ \veta_*\end{smallbmatrix}}
	{\vzero}
	{\begin{smallbmatrix}\mK & \mK_* \\ \mK_*^T & \mK_{**}\end{smallbmatrix}},
	\end{align}
}
Given a novel input $\vx_*$, the posterior distribution of $\veta_*$ is also Gaussian,
	\begin{align}
	p(\veta_*| \mX, \vx_*, \vetabyj) &= \Normalv{\veta_*}{\mK_*^T \mK^{-1} \vetabyj}{\mK_{**} - \mK_*^T \mK^{-1} \mK_*},
	\comments{ \\
	&= \Normalv{\veta_*}
	{\begin{bmatrix}{\vk_*\s{1}}^T {\mK\s{1}}^{-1} \veta\s{1} \\ \vdots \\
		{\vk_*\s{D}}^T {\mK\s{D}}^{-1} \veta\s{D} \end{bmatrix}}
	{\diag \begin{bmatrix} k\s{1}_{**} - {\vk_*\s{1}}^T {\mK\s{1}}^{-1} \vk_*\s{1}
	\\ \vdots \\ k\s{D}_{**} - {\vk_*\s{D}}^T {\mK\s{D}}^{-1} \vk_*\s{D}
	\end{bmatrix}},
	}
	\end{align}
where $\mK = \diag( \mK\s{1}, \cdots, \mK\s{D})$, $\mK_* = \diag( \vk_*\s{1}, \cdots, \vk_*\s{D})$ and $\mK_{**} = \diag(k_{**}\s{1}, \cdots, k_{**}\s{D})$.
$\vk_*\s{j}$ is the j-th cross-covariance matrix with entries $k\s{j}(\vx_*, \vx_i)$, and $k_{**}\s{j} = k\s{j}(\vx_*, \vx_*)$.
\comments{
\\\Rightarrow\quad
	p(\eta_*\s{j}|\mX, \vx_*, \veta\s{j}) &= \Normalv{\eta_*\s{j}}{{\vk_*\s{j}}^T {\mK\s{j}}^{-1} \veta\s{j} }{ k\s{j}_{**} - {\vk_*\s{j}}^T {\mK\s{j}}^{-1} \vk_*\s{j}},
	\end{align}
due to the independence assumption of the latent dimensions.
}
\comments{
or equivalently, due to the independence assumption of the latent dimensions, the novel latent variable  $\eta_*\s{j}$ is distributed as
	\begin{align}
	p(\eta_*\s{j}|\mX, \vx_*, \veta\s{j}) = \Normalv{\eta_*\s{j}}{{\vk_*\s{j}}^T {\mK\s{j}}^{-1} \veta\s{j} }{ k\s{j}_{**} - {\vk_*\s{j}}^T {\mK\s{j}}^{-1} \vk_*\s{j}}
	\end{align}
}
Given the latent vectors, the observation log-likelihood is
	\begin{align}
	\log p(\vybyi|\vtheta(\vetabyi)) = \sum_{i=1}^n \log p(\vy_i | \vtheta(\veta_i))
	= \sum_{i=1}^n \frac{1}{a(\phi)}\left[\vT(\vy_i)^T\vtheta(\veta_i) - b(\vtheta(\veta_i))\right] + \log h(\vy_i).
	\end{align}
Finally, using Bayes' rule, the posterior distribution is
	\begin{align}
	\label{eqn:posteta}
	p(\vetabyj|\mX,\vybyi) = \frac{p(\vybyi|\vtheta(\vetabyi)) p(\vetabyj|\mX)}{p(\vybyi|\mX)},
	\quad
	p(\vybyi|\mX) = \int p(\vybyi|\vtheta(\vetabyi)) p(\vetabyj|\mX) d\vetabyi,
	\end{align}
where $p(\vybyi|\mX)$ is the marginal likelihood.
Finally, given a novel $\vx_*$, the posterior and predictive distributions are obtained by integrating out $\vetabyj$ and $\veta_*$,
	\begin{align}
	\label{eqn:predeta}
	p(\veta_*|\mX,\vx_*,\vybyi) &= \int p(\veta_*|\vetabyj, \mX, \vx_*) p(\vetabyj|\mX,\vybyi)d\vetabyj ,\\
	\label{eqn:predy}
	p(\vy_*|\mX,\vx_*,\vybyi) &= \int p(\vy_*|\vtheta(\veta_*)) p(\veta_*|\mX,\vx_*,\vybyi) d\veta_*.
	\end{align}
As with standard GPs, the hyperparameters of the kernel function and the dispersion can be estimated from the training data by maximizing the marginal likelihood of the training data, $p(\vybyi|\mX)$.
In general, the posterior, marginal, and predictive distributions in (\ref{eqn:posteta}, \ref{eqn:predeta}, \ref{eqn:predy}) cannot be computed in closed-form, and hence approximate inference is necessary.


\section{Approximate Inference for Multivariate GGPMs}
\label{text:approxinf}

In this section, we derive two approximate inference algorithms for M-GGPMs, based on the Taylor \cite{Chan2011CVPR} and Laplace approximations \cite{GPML}.  These two algorithms  use the derivatives of the joint log-likelihood, which in turn can be calculated from the derivatives  of the parameter functions $\mTheta$ of the M-GGPM.  Hence, these algorithms are applicable to {\em any} instance of the M-GGPM.
Other approximate inference methods, such as expectation propagation (EP) \cite{Minka2001},
require integrating $\veta_i$ out of the data likelihood, with respect to a Gaussian distribution.  In general, this integral has no closed-form, and further approximation is required.
EP tends to be susceptible to the quality of these integral approximations, and so finding an efficient  universal integral approximation is a topic of future work.

In general, the posterior distribution is approximated as a Gaussian with  mean $\hat{\vm}$ and covariance $\hat{\mV}$,
	\begin{align}
	p(\vetabyj | \mX, \vybyi) \approx q(\vetabyj | \mX, \vybyi) = \tNormalv{\vetabyj}{\hat{\vm}}{\hat{\mV}}.
	\end{align}
The approximate posterior of $\veta_*$ is then also Gaussian
\comments{
	\begin{align}
	p(\veta_*|\mX,\vx_*,\vybyi) &\approx q(\veta_*|\mX,\vx_*,\vybyi)
	= \int p(\veta_*|\vetabyj, \mX, \vx_*) q(\vetabyj|\mX,\vybyi)d\vetabyj \\
	&= \int \Normalv{\veta_*}{\mK_*^T \mK^{-1}\vetabyj}{\mK_{**}-\mK_*^T \mK^{-1}\mK_*} \Normalv{\vetabyj}{\hat{\vm}}{\hat{\mV}} d\vetabyj \\
	&=\Normalv{\veta_*}{\mK_*^T\mK^{-1} \hat{\vm}}
	{\mK_{**} - \mK_*^T (\mK^{-1} - \mK^{-1} \hat{\mV} \mK^{-1} ) \mK_*}.
	\end{align}
Hence, the approximate posterior of $\veta_*$ is Gaussian with
}
	\begin{align}
	q(\veta_*|\mX,\vx_*,\vybyi) &= \tNormalv{\veta_*}{\hat{\vmu}_*}	{\hat{\mSigma}_*},
	\quad
	\left\{\begin{array}{l}
	\hat{\vmu}_* = \mK_*^T\mK^{-1} \hat{\vm}, \\
	\hat{\mSigma}_* = \mK_{**} - \mK_*^T (\mK^{-1} - \mK^{-1} \hat{\mV} \mK^{-1} ) \mK_*.
	\end{array}\right.
	\end{align}
$\{\hat{\mV}, \hat{\vm}\}$  are determined by the type of approximation, and in many cases take the form
	\begin{align}
	\label{eqn:mggpr_commonform}
	\hat{\mV} = (\mK^{-1} + \mUbyj)^{-1}, \quad
	\hat{\vm} = \hat{\mV} \mUbyj \vtbyj
	=\mK \vzbyj,
	\quad
	\vzbyj = (\mI+\mUbyj\mK)^{-1} \mUbyj \vtbyj
	\end{align}	
for some positive {\em semidefinite} matrix $\mUbyj = \mP^T\mUbyi \mP$, where $\mUbyi=\diag(\mU_1, \cdots, \mU_n)$ is a block diagonal matrix,
and $\mU_i$ approximates the inverse covariance (precision) between latent features for the ith data point.
\comments{
which approximates the covariance between latent features for each datapoint,
	\begin{align}
	\mUbyi = \diag(\mU_1, \cdots, \mU_n),
	\end{align}
where $\mU_i$ is the approximate inverse covariance (precision) between latent features for the ith data point.  We also define the approximate noise covariance matrix as the inverse or pseudo-inverse of $\mU_i$,
	\begin{align}
	\mW_i = \mU_i^{\dagger} \ \ (=\mU_i^{-1})
	\end{align}
as well as the associated matrices
	\begin{align}
	\mWbyi = \diag(\mW_1, \cdots, \mW_n),
	\quad
	\mWbyj = \mP^T\mWbyi \mP.
	\end{align}
The block-diagonal nature implies that this noise is independent between data points.
$\mWbyi$ is this approximate noise matrix ordered by data point, while $\mWbyj$ is the noise matrix ordered by latent dimension.
Note that $\mWbyj$ is not not block diagonal in general, and has structure determined by the permutation matrix $\mP$.
}
Using \refeqn{eqn:mggpr_commonform}, the posterior mean and covariance can be rewritten as
	\begin{align}
	\hat{\vmu}_*
	= \mK^T_* \vzbyj,
	\label{eqn:mggpr_post_mean2U}
	\quad
	\hat{\mSigma}_* = \mK_{**} - \mK_*^T(\mK + \mWbyj)^{-1} \mK_*,
	 \quad \vzbyj = (\mI+\mUbyj\mK)^{-1} \mUbyj \vtbyj
	\end{align}
where $\mWbyj = {\mUbyj}^{-1}$.  If $\mUbyj$ is not invertible, we have assumed that $(\mK+\mWbyj)^{-1}$ can be computed without directly inverting $\mUbyj$ (see \S\ref{text:Ucor} for  details).

\comments{
When $\mU_i$ is invertible, the posterior mean simplifies to
	\begin{align}
	\label{eqn:mggpr_post_mean2}
	\hat{\vmu}_* &= \mK_*^T (\mK + \mWbyj)^{-1} \vtbyj = \mK_*^T\vzbyj,
	\quad
	\vzbyj = (\mK + \mWbyj)^{-1} \vtbyj.
	\end{align}
\TODO{can we show the above also when U is not invertible??}.  However, in general $ (\mI+\mUbyj\mK)^{-1} \mUbyj \neq  (\mK + \mWbyj)^{-1}$ when $\mUbyj$ is not invertible.

Next, we note that
	\begin{align}
	\mK^{-1}-\mK^{-1}\hat{\mV}\mK^{-1}
	&=
	\mK^{-1}-\mK^{-1}(\mK^{-1}+\mUbyj)^{-1}\mK^{-1}
	\\
	&=
	\mK^{-1}-\mK^{-1}(\mK-\mK(\mK+\mWbyj)^{-1}\mK) \mK^{-1}
	= (\mK+\mWbyj)^{-1},
	\end{align}
where we have assumed that, if $\mUbyj$ is not invertible, the matrix $(\mK+\mWbyj)^{-1}$ can be computed directly from $\mUbyj$, rather than inverting $\mUbyj$ (this is discussed later in the implementation section).
The posterior covariance is then
	\begin{align}
	\label{eqn:mggpr_post_var2}
	\hat{\mSigma}_* &= \mK_{**} - \mK_*^T(\mK + \mWbyj)^{-1} \mK_*.
	\end{align}
Note that $\mK$ is block diagonal, but $\mWbyj$ is not block diagonal, in general.    Hence, computing the inverse of $(\mK + \mWbyj)$, in general, will be computationally expensive with complexity $O(n^3D^3)$.  Efficient implementation details for some special cases are discussed in later sections.
}

\subsection{Derivatives of the observation log-likelihood}

The Taylor and Laplace approximations use the derivatives of the data log-likelihood.
The first derivative (as a function of $\veta_i$ and $\vy_i$) is
	\begin{align}
	\vu(\veta_i, \vy_i) &= \tpdd{}{\veta_i} \log p(\vy_i|\vtheta(\veta_i))
	= \tfrac{1}{a(\phi)}\mJ_i^T \left[\vT(\vy_i) - \nabla b(\vtheta(\veta_i))\right],
	\end{align}
where $\mJ_i = \pdd{}{\veta_i^T} \vtheta(\veta_i)$ is the Jacobian of $\vtheta(\veta_i)$.
We also define  the negative Hessian function,
	\begin{align}
	\nonumber
	\mU(\veta_i, \vy_i) &= -
	\tpdd{}{\veta_i} \tpdd{}{\veta_i^T} \log p(\vy_i|\vtheta(\veta_i))
	= \tfrac{1}{a(\phi)}\left[ \mJ_i^T \nabla^2 b(\vtheta(\veta_i)) \mJ_i
	- [\vT(\vy_i) - \nabla b(\vtheta(\veta_i))]^T  \nabla^2\vtheta(\veta_i)
	\right],
	\end{align}
where $\va^T\nabla^2 \vtheta(\veta) = \sum_j a_j \nabla^2 \theta\s{j}(\veta)$.
For the {\em canonical link function}, we have $\mJ_i = \mI$ and $\nabla^2\vtheta(\veta) = \vzero$,
and these derivative functions simplify to
	\begin{align}
	\label{eqn:vu}
	\vu(\veta_i, \vy_i) = \tfrac{1}{a(\phi)}\left[\vT(\vy_i) - \nabla b(\vtheta(\veta_i))\right],
	\quad
	\mU(\veta_i,\vy_i) = \tfrac{1}{a(\phi)}\nabla^2 b(\vtheta(\veta_i)).
	\end{align}
\comments{
	\begin{align}	
	\mW(\veta_i, \vy_i) &= a(\phi)\left[\nabla^2 b(\vtheta(\veta_i)) \right]^{\dagger}.
	\end{align}
}
For  convenience, we will use the shorthand $\vu_i = \vu(\veta_i,\vy_i)$ and $\mU_i = \mU(\veta_i,\vy_i)$.
Finally, we define $\mW_i = \mU_i^{-1}$ when $\mU_i$ is invertible, or $\mW_i = \mU_i^{\dagger}$
(the pseudo-inverse) otherwise.  Note that in both cases, $\vu_i = \mU_i\mW_i \vu_i$, since in the latter-case $\vu_i$ must be in the column-space of $\mU_i$ by construction.

\comments{
	
Consider the case when the likelihood is constant in a hyperplane of the solution space, i.e.
	\begin{align}
	\log p(\vy_i|\vtheta(\veta_i)) = \log p(\vy_i|\vtheta(\veta_i + \delta\va))
	\end{align}
for some vector $\va\in \real^D$ and $\delta\in\real$.  For example, this is true for the multinomial with $\va=\vone$.
This condition implies that the derivative function is orthogonal to $\va$, and also that the null-space of $\mU_i$ is $\va$, i.e.,
	\begin{align}
	\va^T\vu_i = 0, \quad\quad
	\mU_i\va = \vzero.
	\end{align}
Hence, $\mU_i$ is not invertible.  From the above condition, we have that $\vu_i$ is in the column-space of $\mU_i$, i.e., for some $\bar{\vu}_i$,
	\begin{align}
	\label{eqn:deg_vu}
	\vu_i &= \mU_i \bar{\vu}_i
	= \mU_i \mW_i \mU_i \bar{\vu}_i
	= \mU_i \mW_i \vu_i,
	\end{align}
which follows from the pseudo-inverse property $\mU_i \mW_i\mU_i = \mU_i$.
}

\subsection{Taylor Approximation}

We next derive the Taylor approximation \cite{Chan2011CVPR} for M-GGPM inference.
%
In particular,
the log-likelihood of $\vy_i$ 
 is approximated with
a 2nd-order Taylor expansion around the point $\tveta_i$,
	\begin{align}
	\log p(\vy_i|\vtheta(\veta_i))
	&\approx
	v_i + \tvu_i^T(\veta_i - \tveta_i) - \tfrac{1}{2}(\veta_i-\tveta_i)^T\tmU_i (\veta_i-\tveta_i)+ \log h(\vy_i)
	\label{eqn:tayloreachy}
	\end{align}
where $\tvu_i = \vu(\tveta_i,\vy_i)$, $\tmU_i = \mU(\tveta_i,\vy_i)$, and
$v_i = \tfrac{1}{a(\phi)}\left[\vtheta(\tveta_i)^T \vT(\vy_i) - b(\vtheta(\tveta_i))\right]$.
Using \refeqn{eqn:tayloreachy}, the joint distribution $p(\vybyi,\vetabyj|\mX)$ is approximated as
	\begin{align}
	\begin{split}
	\log q(\vybyi,\vetabyj|\mX)
	& =
	\log p(\vybyi|\vtheta(\tvetabyi))
	- \tfrac{1}{2} \log \detbar{\mK}
	-\tfrac{1}{2} \|\vetabyj - \mA^{-1}\tmUbyj \tvtbyj \|^2_{\mA^{-1}}
	\\	&\quad\quad
	-\tfrac{1}{2}(\tvtbyj)^T (\mI+\tmUbyj\mK)^{-1}\tmUbyj \tvtbyj
	+ \tfrac{1}{2}\tvubyi^T\tmWbyi\tvubyi  - \tfrac{nD}{2}\log 2\pi,
	\end{split}
	\label{eqn:jointapprox}
	\end{align}
where $\mA = \mK^{-1}+\tmUbyj$,
$\tvtbyj = \tvetabyj + \tmWbyj\tvubyj$ is the ``target'' vector,
and $\tmW_i = \mW(\tveta_i,\vy_i)$.
%
Dropping the terms of \refeqn{eqn:jointapprox} that do not depend on $\vetabyj$, the Taylor approximation of the posterior $p(\vetabyj|\mX,\vybyi)$ is Gaussian of the form in \refeqn{eqn:mggpr_commonform}, with mean and covariance
	\begin{align}
	\hat{\mV} = ( \mK^{-1} + \tmUbyj )^{-1},
	\quad
	\hat{\vm} = \hat{\mV}\tmUbyj\tvtbyj
	=\mK \tvzbyj,
	\quad
	\tvzbyj = (\mI+\tmUbyj\mK)^{-1} \tmUbyj \tvtbyj
	\label{eqn:cfa:post}
	\end{align}
\comments{
These are of the form in \refeqn{eqn:mggpr_commonform}, and hence the posterior is given in
\refeqn{eqn:mggpr_post_mean2}, \refeqn{eqn:mggpr_post_mean2U}, and \refeqn{eqn:mggpr_post_var2} with
	\begin{align}
	\mUbyi &= \diag(\tmU_1 ,\cdots ,\tmU_n), \\
	\mWbyi &= \diag(\tmW_1, \cdots, \tmW_n), \\
	\vtbyj &= \tvetabyj + \tmWbyj\tvubyj = \tvetabyj + \mP^T\tmWbyi\mP\tvubyj
	= \mP (\tvetabyi + \tmWbyi\tvubyi), \\
	\vzbyj &= (\mI+\tmUbyj\mK)^{-1}\tmUbyj \tvtbyj
	\end{align}
}
where $\tvtbyj$ and $\tmUbyj$ are determined by the expansion point $\tveta_i$.  In general, $\tveta_i$ can be selected as a function of the observation $\vy_i$.
One interesting case is when expansion point is $\tveta_i = \vg(\vT(\vy_i))$, i.e., we assume the observation is close to the mean.  The target and inverse covariance simplify to
	\begin{align}
	\tvt_i 
	= \vg(\vT(\vy_i)),
	\quad
	\tmU_i &= \tfrac{1}{a(\phi)} \mJ_i^T \nabla^2 b(\vtheta(\tveta_i)) \mJ_i.
	\end{align}
Finally, the marginal likelihood is obtained by integrating out $\vetabyj$ from the joint likelihood in \refeqn{eqn:jointapprox},
	\begin{align}
	\log q(\vybyi|\mX)
	&=
	-\tfrac{1}{2}(\tvtbyj)^T (\mI+\tmUbyj\mK)^{-1}\tmUbyj \tvtbyj
	- \tfrac{1}{2}\log |\mI + \tmUbyj\mK|
	+  r(\phi),
	\label{eqn:mggpm_taylor_marginal}
	\end{align}
where $r(\phi) = \sum_i \log p(\vy_i|\vtheta(\tveta_i)) + \tfrac{1}{2}\tvu_i^T\tmW_i\tvu_i$ depends only on the dispersion.
As a non-iterative approximation, Taylor inference is more efficient than other iterative methods (e.g. Laplace, or EP).

\comments{
where $r(\phi,\mPhi)$  depends only on  the dispersion hyperparameters
	\begin{align}
	r(\phi, \mPhi) &= \log p(\vybyi|\vtheta(\tvetabyi))
	+ \frac{1}{2}\tvubyi^T\tmWbyi\tvubyi
	  = \sum_{i} r_i(\phi, \mPhi),
	 \\
	 r_i(\phi, \mPhi) &= \log p(\vy_i|\vtheta(\tveta_i))
	+ \frac{1}{2}\tvu_i^T\tmW_i\tvu_i .
	\end{align}
}

\subsection{Laplace Approximation}

The Laplace approximation is a second-order Taylor approximation at the mode of the posterior, i.e.
	\begin{align}
	\hvetabyj = \argmax_{\vetabyj} \log p(\vetabyj|\mX,\vybyi)
	= \argmax_{\vetabyj} \log p(\vybyi|\vtheta(\vetabyi)) +\log p(\vetabyj|\mX).
	\end{align}
The maximum $\hvetabyj$ can be found iteratively using the Newton-Raphson algorithm.  In particular, given an initial estimate $\hvetabyj$, a new estimate is found iteratively according to
	\begin{align}
	\hvetabyj_{\mathrm{new}} 
	&= \mK \hvzbyj, \quad
	\hvzbyj = (\mI + \hmUbyj\mK)^{-1}(\hvubyj +\hmUbyj\hvetabyj )
	\end{align}
where $\hmUbyj$ and $\hvubyj$ are evaluated at the current $\hvetabyj$.
The Newton iteration can also be rewritten as
	\begin{align}
	\hvetabyj_{\mathrm{new}}
	&= \mK \hvzbyj, \quad
	\hvzbyj = (\mI + \hmUbyj\mK)^{-1}\hmUbyj \hvtbyj,
	\quad
	\hvtbyj = \hmWbyj\hvubyj + \hvetabyj.
	\label{eqn:laplace-update}
	\end{align}
\comments{
Note that
	\begin{align}
	\hvubyj +\hmUbyj\hvetabyj &= \hmUbyj\hmWbyj\hvubyj +\hmUbyj\hvetabyj \\
	&= \hmUbyj(\hmWbyj\hvubyj + \hvetabyj) \\
	\Rightarrow\hvubyj +\hmUbyj\hvetabyj &= \hmUbyj\hvtbyj,
	\quad
	 \hvtbyj = \hmWbyj\hvubyj + \hvetabyj,
	\end{align}
and thus,
}
Given the mode, the approximate posterior is Gaussian of the form in \refeqn{eqn:mggpr_commonform}, with parameters
	\begin{align}
	\hat{\mV} &= (\mK^{-1}+\hmUbyj)^{-1}, \quad
	\hat{\vm} = \hvetabyj =\mK \hvzbyj = \hat{\mV}\hmUbyj\hvtbyj.
	\label{eqn:laplace-post}
	\end{align}
%
%
To approximate the marginal log-likelihood, a second-order Taylor approximation is applied to the joint log-likelihood at $\hvetabyj$, and $\vetabyj$ is integrated out, yielding
\comments{
	\begin{align}
	\log p(\vybyi, \vetabyj|\mX) \approx  \log p(\vybyi|\vtheta(\hvetabyi)) + \log p(\hvetabyj|\mX)
	- \tfrac{1}{2} \|\vetabyj - \hvetabyj\|^2_{\hat{\mV}}
	\end{align}
Exponentiating and integrating out $\vetabyj$ yields an approximation to the marginal log-likelihood,
}
	\begin{align}
	\log q(\vybyi|\mX)
	&= \log p(\vybyi|\vtheta(\hvetabyi)) - \tfrac{1}{2}(\hvetabyj)^T\mK^{-1}\hvetabyj
	-\tfrac{1}{2} \log |\mI+\hmUbyj \mK|.
	\end{align}
Comparing \refeqn{eqn:cfa:post} and \refeqn{eqn:laplace-update} shows that the Taylor approximation can be interpreted as a one-iteration Laplace approximation starting at the expansion point $\tvetabyi$.  Hence, we expect the Taylor approximation to give reasonable results when the expansion point is chosen wisely.

\subsection{Efficient Implementation}
\label{text:Ucor}


For the Laplace and Taylor approximations, the main computational bottleneck is in inverting $(\mI + \mUbyj \mK)$.
In general, $\mK$ is block diagonal and $\mUbyj$ is not block diagonal. Hence, naively calculating this inverse has  complexity $O(n^3D^3)$.
The efficiency can be  improved if the negative Hessian function takes a particular form, namely
	\begin{align}
	\label{eqn:Ucor}
	\mU_i = \mGamma_i + a_i \vomega_i\vomega_i^T,
	\quad
	\mUbyi = \mGammabyi+\mOmegabyi\mAbyi\mOmegabyi^T,
	\end{align}
where $\mGamma_i = \diag(\vgamma_i)$ is a diagonal matrix and $\vgamma_i\in\real^D$, $\vomega_i\in\real^D$ models correlation between latent dimensions in the Hessian, and $a_i\in\real$.
$\mGammabyi$, $\mOmegabyi$, and $\mA$ are the corresponding (block) diagonal matrices.
%
%
Reordering by latent dimension to get $\mUbyj$, and applying the matrix inversion lemma,
	\begin{align}
	\mUbyj =
	 \mGammabyj + \mOmegabyj \mA {\mOmegabyj}^T,
	\quad
	{\mUbyj}^{-1} 
	=
	\mLbyj - \mMbyj(\mA^{-1} + \mMbyi^T\mGammabyi\mMbyi)^{-1}{\mMbyj}^T,
	\label{eqn:corr_Uinv}
	\end{align}
where
$\mGammabyj = \mP^T\mGammabyi \mP$ is the reordered diagonal matrix, and
$\mOmegabyj = \mP^T \mOmegabyi$ is a stack of diagonal matrices with
$\mOmega\s{j} = \diag(\vomega\s{j})$.
For the inverse,
$\mLbyj = {\mGammabyj}^{-1}$, and
$\mMbyj = \mLbyj \mOmegabyj$ is a stack of diagonal matrices where
$\mM\s{j} = \diag(\mL\s{j}\vomega\s{j})$.
%
\comments{
	\begin{align}
	\mMbyj = \mP^T \mMbyi = \mP^T
	\begin{bmatrix}
	\vm_1 & 0 & 0 \\
	0 & \ddots & 0 \\
	0 & 0 & \vm_n
	\end{bmatrix}
	=
	\begin{bmatrix}
	\mM\s{1} \\ \vdots \\ \mM\s{D}
	\end{bmatrix},
	\quad
	\mM\s{j} =
	\begin{bmatrix}
	m_1\s{j} & 0 & 0 \\
	0 & \ddots & 0\\
	0 & 0 & m_n\s{j}
	\end{bmatrix},
	\end{align}
where $m_i\s{j}$ is the jth element of $\vm_i$.	

A special case of the correlated Hessian model is when $\mU_i$ is singular, i.e.
	\begin{align}
	a_i^{-1}+\vomega_i^T\mL_i\vomega_i = 0.
	\end{align}
For example, this occurs for the multinomial when $\vgamma_i = \vomega_i/N$, $a_i=-1/N$, and $\vone^T\vomega_i=1$.
}
\comments{
	\begin{align}
	\vone^T(\mGamma_i+a_i\vomega_i\vomega_i^T)\vone = 0 \quad
	\Rightarrow \quad \vone^T\vgamma_i = -a_i(\vone^T\vomega_i)^2.
	\end{align}
For example, with the multinomial, $\vgamma_i = \vomega/N$, $a_i=-1/N$, and $\vone^T\vomega_i=1$.
}

Next, we look at efficiently computing $(\mK+\mWbyj)^{-1}$ using only $\mUbyj$.
Substituting with \refeqn{eqn:corr_Uinv} and applying the matrix inversion lemma yields,
	\begin{align}
	(\mK+\mWbyj)^{-1}
	&= \mB^{-1} - \mB^{-1}\mMbyj\hat{\mA}{\mMbyj}^T\mB^{-1} ,
	\label{eqn:invKW}
	\end{align}
where  $\mB = \mK+\mLbyj$ is a block diagonal matrix,  and
	$\hat{\mA} = (-\mA^{-1} - \mMbyi^T\mGammabyi\mMbyi + {\mMbyj}^T\mB^{-1}\mMbyj)^{-1}$.
Next, we apply the matrix inversion lemma to $(\mK^{-1}+\mUbyj)^{-1}$ and use \refeqn{eqn:invKW},
	\begin{align}
	(\mK^{-1}+\mUbyj)^{-1}
	&= \mK - \mK(\mK+\mWbyj)^{-1}\mK 
	=\mK\mB^{-1}\mLbyj+ \mK\mB^{-1}\mMbyj\hat{\mA}{\mMbyj}^T\mB^{-1}\mK
	\label{eqn:invKU}
\\
\Rightarrow & \quad
	(\mI+\mUbyj\mK)^{-1} 
	=\mB^{-1}\mLbyj + \mB^{-1}\mMbyj\hat{\mA}{\mMbyj}^T\mB^{-1}\mK.
	\label{eqn:invIUK}
	\end{align}
Note that in (\ref{eqn:invKW}, \ref{eqn:invKU}, \ref{eqn:invIUK}),
the inversion of a $nD\times nD$ matrix has been reduced to inverting $\mB$, a block-diagonal matrix with $n\times n$ block entries, and the $n \times n$ matrix $\hat{\mA}$.
Hence, the overall complexity is reduced to $O(D n^3)$. 
In addition, these inverses can be calculated without explicitly inverting $\mU_i$, thus avoiding problems when $\mU_i$ is singular.  A similar trick was used to calculate (\ref{eqn:invKW}, \ref{eqn:invKU}) for multi-class GPC in \cite{Williams1998GPC,GPML}, and here we have derived a more general case.
Finally, the log-determinant can also be computed efficiently by applying the matrix determinant lemma,
	$|\mI+\mUbyj\mK|   = |\hat{\mA}^{-1}| |\mGammabyj| |\mB ||\mA|$,
which reduces to calculating  determinants of $n \times n$ matrices.

\section{Examples and Experiments}
\label{text:examples}

In this section, we present several models derived from the M-GGPM, along with  experiments.

\subsection{Multinomial Distribution}

We first consider the $d$-dim. multinomial distribution over normalized counts $\vy \in \{\tfrac{0}{N},\cdots,\tfrac{N}{N}\}^d$,
	\begin{align}
	p(\vy|\vpi,N) = \tbinom{N}{N\vy} \prod_{j=1}^d \pi_j^{Ny_j}
	\end{align}
where $0\leq \vpi \leq 1$ is the vector of  probabilities, and $N$ is the number of trials.
Setting $\vtheta = \log \vpi$, the exponential family parameters are
	\begin{align}
	\vT(\vy) = \vy,
	\quad
	b(\vtheta) = \log \vone^Te^{\vtheta},
	\quad
	h(\vy,\phi) = \tbinom{N}{N\vy},
	\quad
	a(\phi) = 1/N.
	\end{align}
Assuming the canonical link function, the derivative functions are
	\begin{align}
	\vu_i = N (\vy_i-\vpi_i) ,
	\quad
	\mU_i = N (\diag(\vpi_i) - \vpi_i\vpi_i^T),
	\quad
	\mW_i = 
	\tfrac{1}{N}\left[\diag(\vpi_i)^{-1} - c\vone\vone^T\right],
	\end{align}
where $\vpi_i = \tfrac{e^{\veta_i}}{\vone^Te^{\veta_i}}$.
The Hessian $\mU_i$ can be written in the efficient form of \refeqn{eqn:Ucor}, but is degenerate.
$\mW_i$ is the pseudo-inverse of $\mU_i$ when $c=1$, and a generalized-inverse otherwise.
For $N=1$, the multinomial-GGPM is identical to multi-class GP classification with the soft-max function \cite{Williams1998GPC, GPML}.

We next consider the Taylor approximation for multinomial-GGPM.  The target vector for $\tveta_i$ is
$\tvt_i =  \tveta_i + \diag(\tilde{\vpi}_i)^{-1}\vy_i - \vone$.
Ignoring the class information and setting the expansion point to $\tveta_i = \vzero$,
we obtain the Taylor target and Hessian matrices
	\begin{align}
	\tvt_i = d \vy_i - \vone,
	\quad
	\tmU_i = N (\tfrac{1}{d}\mI - \tfrac{1}{d^2}\vone\vone^T) ,
	\quad
	\tmW_i = \tfrac{1}{N} (d\mI - c\vone\vone^T), \quad
	\end{align}
If the class counts are only $\{0,1\}$, then
each observation is  $\vy_i=\ve_j$, and
the corresponding target $\vt_i$ is a vector with $d-1$ in index $j$, and $-1$ otherwise.  Hence, for each latent dimension, the target $\vt\s{j}$ contains $d-1$ for all examples in class $j$ (positive classes), and $-1$ otherwise (negative classes).  Therefore, the Taylor approximation for multi-class GPC can be interpreted as {\em implementing a set of 1-vs-all label regression classifiers}.
With $\tmW_i$ further approximated as a diagonal noise matrix (by setting $c=0$), the Taylor approximation yields $d$ {\em independent} GP label regression classifiers,
where $N$ act as a hyperparameter similar to $\sigma^2$ in GPR.
This interpretation of 1-vs-all GP label regression as a Taylor approximation of multi-class GPC offers a theoretical justification for why label regression tends to work work well in practice,
e.g. in \cite{Kapoor2010objects}.

Finally, using a single latent dimension ($D=1)$, GP ordinal regression \cite{Chu2005GPOR} is obtained by setting $\theta\s{j}(\eta) = \log [\Phi(b_j - \eta) - \Phi(b_{j-1} - \eta)]$, where $\Phi(x)$ is the cumulative Gaussian function and $b_j$ are the interval mappings that divide $\eta$ into ordinal bins.

\subsection{Von-Mises Distribution}

The Von-Mises distribution is a circular distribution over radian angles $y \in \real$  given by
	\begin{align}
	p(y|\mu,\kappa) = \tfrac{1}{2\pi I_0(\kappa)} e^{\kappa \cos(y-\mu)},
	\quad
	I_n(\kappa) = \tfrac{1}{\pi} \int_0^{\pi} e^{\kappa \cos y} \cos(ny) dy,
	\end{align}
where $\mu \in \real$ is the circular mean,  $\kappa>0$ is the concentration, and	
$I_n(\kappa)$
is the modified Bessel function of the first kind. 
Setting $\vtheta = \begin{smallbmatrix}\kappa \cos \mu \\ \kappa\sin\mu\end{smallbmatrix}$,
the exponential family parameters are
	\begin{align}
	\vT(y) = \begin{smallbmatrix} \cos y \\ \sin y\end{smallbmatrix},
	\quad
	b(\vtheta) = \log I_0(\norm{\vtheta}),
	\quad
	h(y) = \tfrac{1}{2\pi},
	\quad
	a(\phi) = 1.
	\end{align}
Assuming the canonical link function, the derivative functions are
	\begin{align}
	\vu_i &=  \vT(y_i) - c_1(\kappa_i) \veta_i,
	\  \
	\mU_i &=  c_1(\kappa_i) \mI + \tfrac{c_2(\kappa_i)}{\kappa_i^2} \veta_i\veta_i^T ,
	\ \
	\mW_i &= 	\tfrac{1}{c_1(\kappa_i)} [ \mI - \tfrac{c_3(\kappa_i)}{\kappa_i^2}\veta_i\veta_i^T ],
	\end{align}
where $\kappa_i = \norm{\veta_i}$ is the length of $\veta_i$, and the constants are
	$c_1(\kappa) = \tfrac{I_1(\kappa)}{I_0(\kappa) \kappa}$,
	$c_2(\kappa) 
	= 1 - \kappa^2 c_1(\kappa)^2 - 2c_1(\kappa)$,
	$c_3(\kappa) = \tfrac{c_2(\kappa)}{c_1(\kappa)+c_2(\kappa)}$.
Note that the latent vector $\veta$ models {\em both} the concentration (via the length) and the circular mean (via the angle) of the Von-Mises distribution.
In addition, the Hessian can be written in the form of \refeqn{eqn:Ucor}, allowing for a more efficient implementation.
For the Taylor approximation, we set the expansion point to the observed direction $\tveta_i = \vT(y_i)$, resulting in 
	\begin{align}
	\tvt_i  
	= \tfrac{1+c_2}{c_1+c_2}\vT(y_i),
	\quad
	\tmU_i = c_1\mI + c_2\vT(y_i)\vT(y_i)^T,
	\quad
	\tmW_i = \tfrac{1}{c_1}\left[\mI - c_3 \vT(y_i) \vT(y_i)^T\right]
	\end{align}
where $c_1 = c_1(1)$, $c_2 = c_2(1)$ and $c_3 = c_3(1)$.  The targets are scaled versions of $\vT(y_i)$, while the noise adds additional variance in the direction of these targets.

\begin{figure}
\hspace{-0.1in}
\begin{tabular}{c@{}c@{}c}
\footnotesize{\ \ \ \ (a)} & \footnotesize{\ \ (b)} & \footnotesize (c) \vspace{-0.1in}\\
\includegraphics[width=0.30\linewidth]{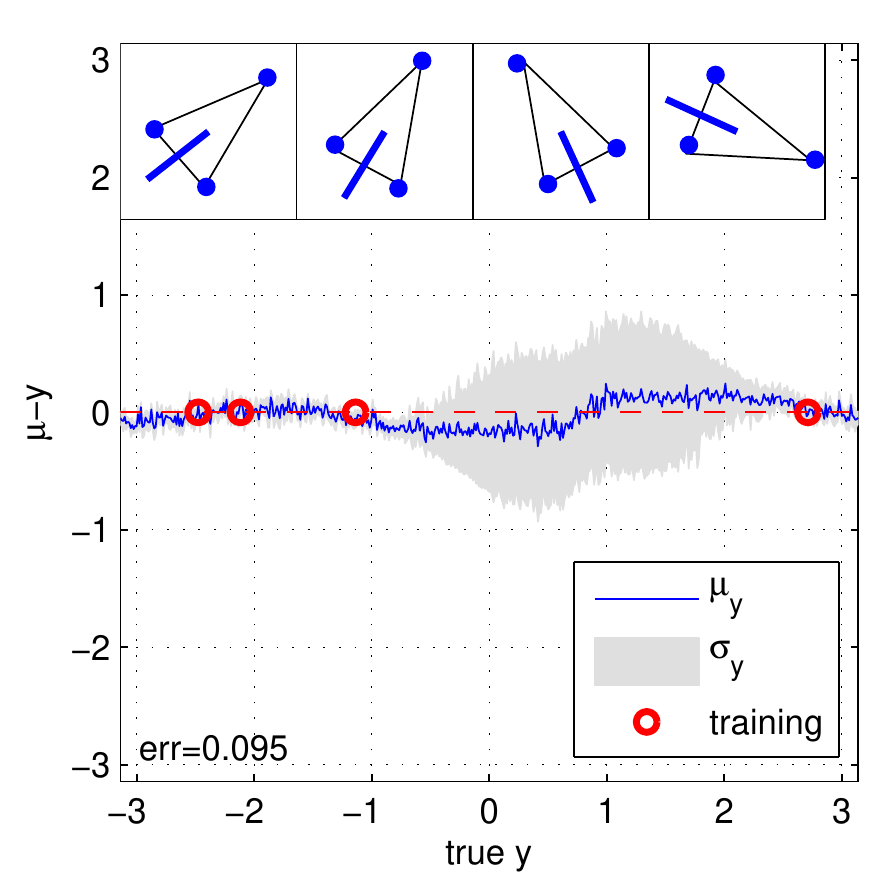}  &
\includegraphics[width=0.34\linewidth]{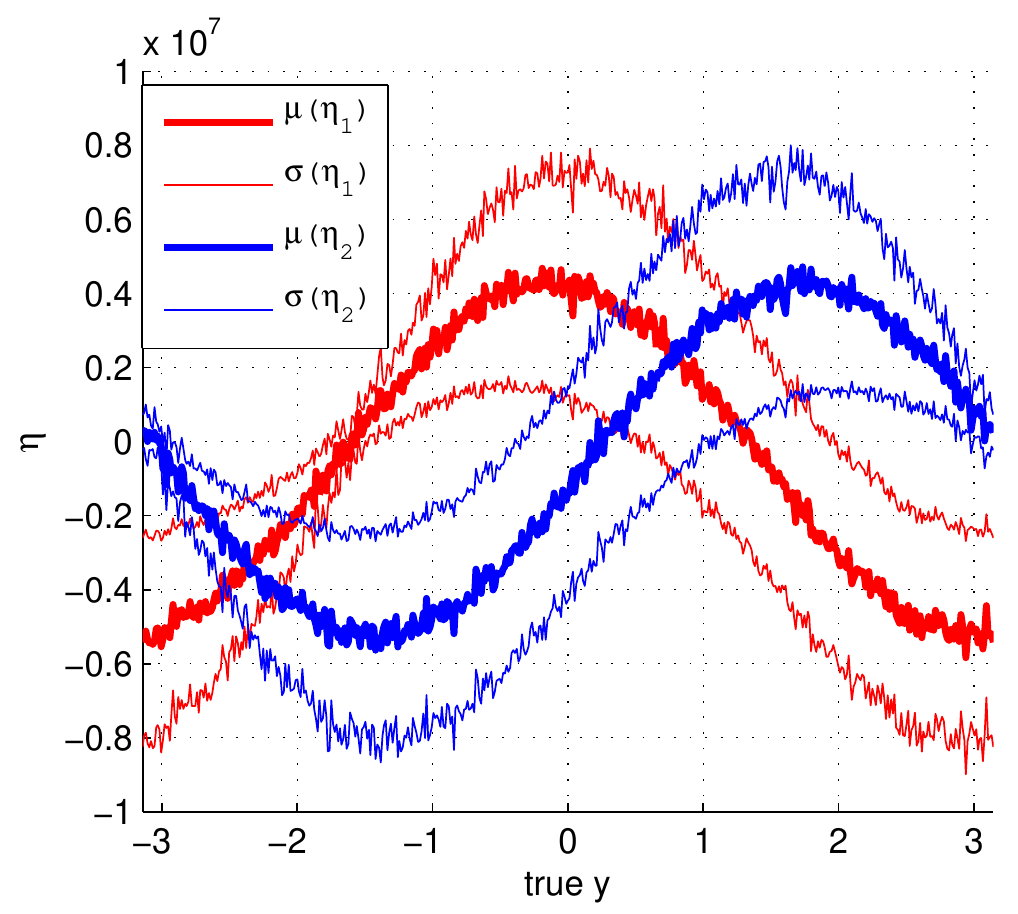}  &
\includegraphics[width=0.37\linewidth]{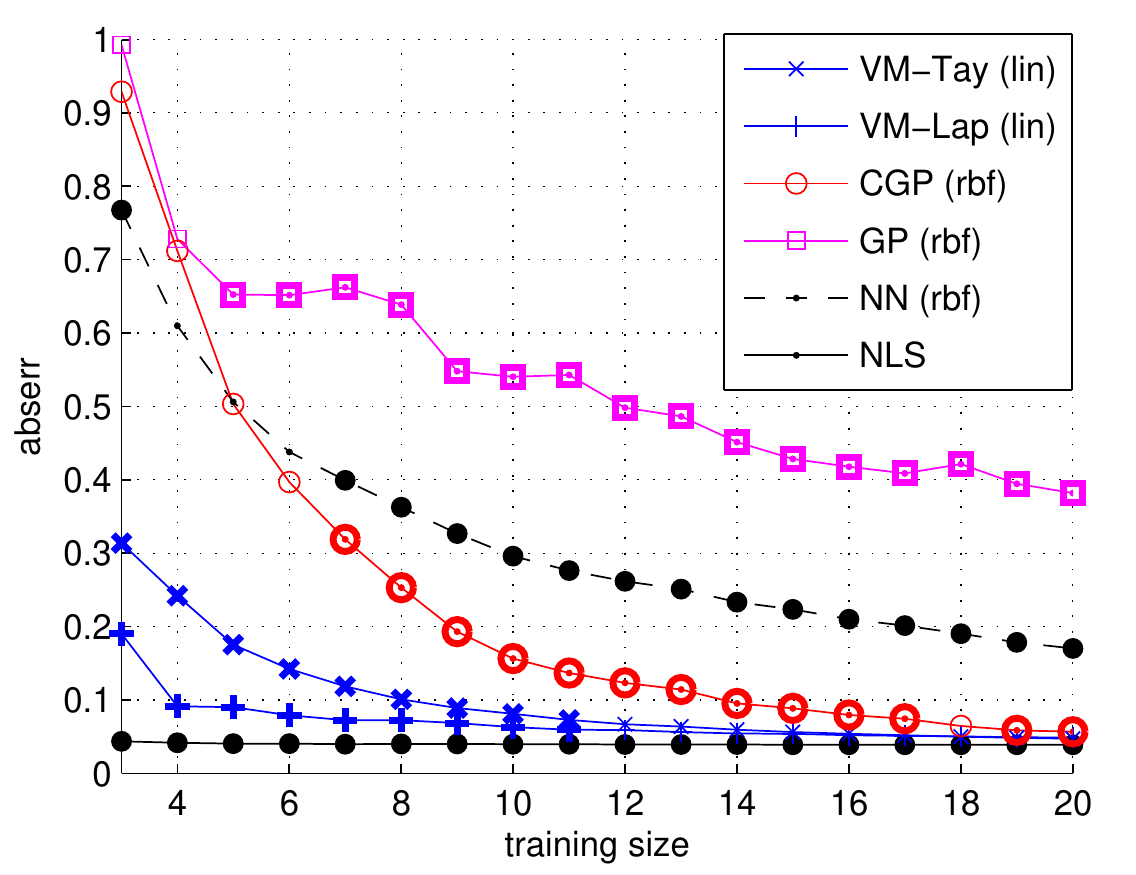}
\end{tabular}
\vspace{-0.2in}
\caption{\footnotesize Angle regression with the Von-Mises GGPM: a) 4 training shapes and headings, and the error between the predictive distribution and the true angle; b) the corresponding 2-d latent vectors for each test angle; c) the average absolute error versus number of training shapes.  Bold markers indicate when the performance difference is significant (paired t-test between two neighboring methods yields $p<0.001$).}
\label{fig:anglesreg}
\vspace{-0.2in}
\end{figure}

We experiment with the Von-Mises GGPM for predicting the orientation of a shape, given  noisy observations of the corners. 
A triangle is formed by randomly sampling three points from a 2-d unit Gaussian distribution, and an arbitrary ground-truth heading is sampled from a uniform distribution over $[-\pi, \pi)$.
Noisy observations of the shape are generated by randomly rotating the triangle, followed by adding Gaussian noise ($\sigma=0.1$) to the points.
%
A Von-Mises GGPM is learned using either the Taylor or Laplace approximations (VM-Tay and VM-Lap), where each input vector is the concatenation of the triangle points (the point ordering is known), and the output value is the observed heading.  The hyperparameters are estimated by maximizing marginal likelihood.  Next, given a novel (noisy) triangle, the heading is estimated as the predictive mean.  Figures \ref{fig:anglesreg}a and \ref{fig:anglesreg}b show examples of the training shapes, and the corresponding  predictive and latent distributions.

The absolute error of the prediction was calculated by averaging over 50 trials with 100 test triangles per trial.  For comparison, we also report the errors for standard GPR (GP),  the quadratically-constrained GP\footnote{Of the 2 ambiguous predictions produced by CGP, we select the prediction closest to the nearest-neighbor prediction.} (CGP) \cite{Salzmann2010CGP}, and nearest-neighbors (NN).
Figure \ref{fig:anglesreg}c presents the prediction error for different training sizes, using the best kernel (linear or RBF) for each method.
Both versions of the Von-Mises GGPM perform significantly better than GP, CGP, and NN.  In addition, while the Laplace approximation is more accurate than the Taylor approximation when the training size is small, the difference in performance is not significant when the training size is increased.
Finally, we compare to non-linear least square (NLS), which is an optimal solution that directly uses the {\em functional form} of the rotation matrix.
The Von-Mises GGPM performs similarly to NLS, especially for larger training sizes, even though the M-GGPM does not use any specific assumptions about the rotation model.

\subsection{Dirichlet Distribution}
The Dirichlet distribution is a distribution over the $d-1$ simplex (i.e., $d$-dim. probability vectors),
	\begin{align}
	p(\vy|\valpha) = \tfrac{1}{B(\valpha)}\prod_{j=1}^d y_j^{\alpha_j-1},
	\quad
	B(\valpha) = \tfrac{\prod_{j=1}^d \Gamma(\alpha_j)}{\Gamma(\vone^T\valpha)},
	\quad
	\vone^T\vy  = 1, \quad \vy>0,
	\end{align}
where $\valpha > 0$, and $B(\valpha)$ is the Beta function.  Setting $\vtheta = \valpha$, the exponential family parameters are
	\begin{align}
	\vT(\vy) = \log \vy,
	\quad
	b(\vtheta) = \log B(\vtheta),
	\quad
	h(\vy,\phi) = \tfrac{1}{\prod_{j=1}^d y_j},
	\quad
	a(\phi) = 1.
	\end{align}
Because $\vtheta=\valpha$,  there is an implicit restriction that $\vtheta>0$. 
Hence, we select $\vtheta(\veta)$ as the logistic error function, which ensures $\vtheta$ is positive, while modeling a linear trend between between $\veta$ and $\vtheta$,
	\begin{align}
	\vtheta(\veta_i) = \log( e^{\veta_i} + 1 ),
	\quad
	\mJ_i = \diag( \tfrac{e^{\veta_i}}{1+e^{\veta_i}} ),
	\quad
	\nabla^2 \theta\s{j}(\veta_i) = \diag( \tfrac{e^{\veta_i}}{(1+e^{\veta_i})^2}).
	\end{align}
Using this parameterization, the derivative functions are
    \begin{align}
    \vu_i = \mJ_i^T[\log(\vy_i) - \nabla B(\vtheta(\veta_i))],
    \quad
    \mU_i = \mGamma_i + a_i\vomega_i\vomega_i^T
    \quad
    \end{align}
where $\mGamma_i = \diag(\psi_1(\vtheta(\veta_i)) - \log\vy_i + \psi_0(\vtheta(\veta_i)) - \psi_0(\vone^T\vtheta(\veta_i))\vone)$, $\vomega_i = \diag(\mJ_i)$, $a_i = -\psi_1(\vone^T\vtheta(\veta_i))$, 
and $\psi_n(x) = \tpdd{^{n+1}}{x^{n+1}} \log \Gamma(x)$ is the polygamma function of order $n$. 
The Hessian is of the form in \refeqn{eqn:Ucor}, allowing for efficient inference.  For the Taylor approximation, we set the expansion point to be the observation, $\tveta_i = \vy_i$. 

\begin{figure}[t]
\includegraphics[width=1\linewidth]{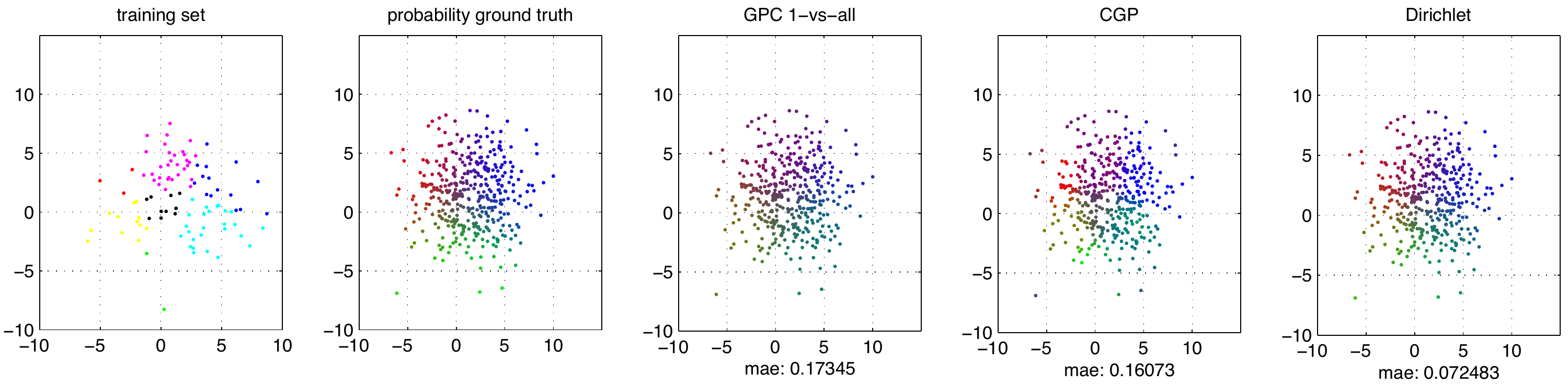}
\vspace{-0.3in}
\caption{\footnotesize 
Multinomial regression: the training targets are quantized probability vectors.  Predicted probability values are represented with RGB colors.
}
\label{fig:labelregression}
\vspace{-0.2in}
\end{figure}

We conduct two experiments with the Dirichlet-GGPM. 
The first task is to recover a probability distribution from a noisy (quantized) distribution.
Random samples are drawn around three 2-d Gaussians, and a vector is formed with the posterior probability of the sample belonging to each Gaussian.
Next, the probabilities are quantized by thresholding the probability vector at $0.2$, and re-normalizing.
The Dirichlet-GGPM is used to recover the underlying probability distribution of test points given a training set of points and noisy probability vectors.
CGP and binary GP classification (GPC) are also trained for comparison.
All models used the RBF kernel, and hyperparameters were estimated by maximizing the marginal likelihood.  
Figure \ref{fig:labelregression} presents the training set and the predicted probability maps.  
Dirichlet-GGPM is able to recover the underlying probability vectors, with mean absolute error (MAE) of 0.072.  On the other hand, 
CGP only ensures the linear constraint of the probability vector, but does not enforce that the entries are probabilities.  Values not in $[0, 1]$ must be truncated, resulting in artifacts in prediction (MAE of 0.16).
Finally, because GPC assumes independent output dimensions, the 1-vs-all class probabilities must be normalized, resulting in poorly calibrated probability distributions (MAE of 0.17)
%

\begin{figure}[b]
\vspace{-0.2in}
\centering
\includegraphics[width=0.7\linewidth]{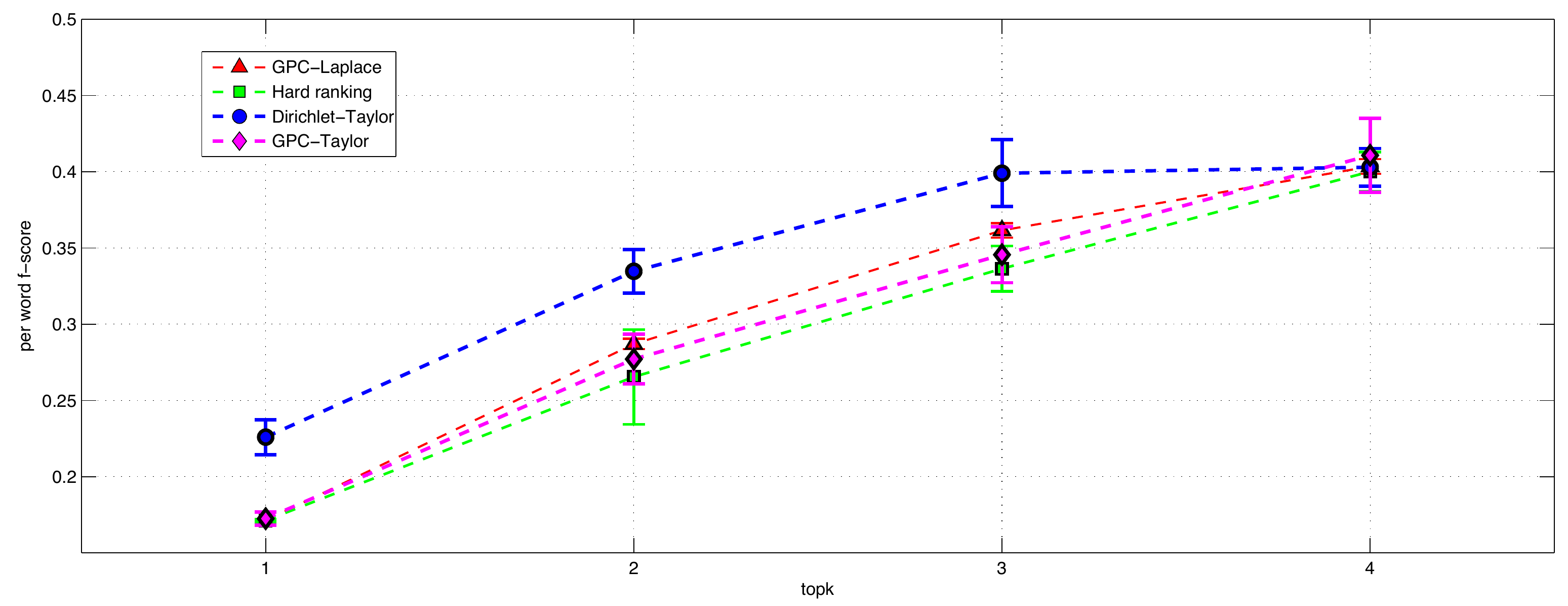}
\vspace{-0.2in}
\caption{\footnotesize 
Image annotation results using Dirichlet-GGPM.
}
\label{fig:imageannotation}
\vspace{-0.2in}
\end{figure}

We also apply Dirichlet-GGPM to an image annotation task, for mapping image features (HUV color hisograms) to probabilities of semantic concepts.
This is similar to the synthetic experiment above, in that the training labels only indicate the presence/absence of the concept, and not the actual probability.  Given a test image, the Dirichlet-GGPM predicts the probabilities of the semantic concepts, which is then used for ranked annotation.
%
%
%
%
The dataset consists of 482 images that are labeled with the 10 most frequent annotations of the Corel5k  dataset \cite{Feng2004mbrm}.  300 images were used for training, with the remainder reserved for testing. 
The Dirichlet-GGPM was trained with the RBF kernel, and compared to two baseline models, binary GPC (treating each label as a 1-vs-all classification task), and annotations based on the most frequent labels (``hard ranking'').
Figure \ref{fig:imageannotation} show the average per-word F-score for annotation with $k$ labels, with results averaged over 10 random splits of the training/test data.
Dirichlet-GGPM outperforms the other models significantly on the top-ranked annotations.





\comments{
\subsubsection*{Acknowledgments}
Ack!
GPML code.
grants (startup, GRF)
}

\newpage
\bibliographystyle{ieeetr}
\bibliography{../../../biblio/abc_all,../../../biblio/GGPRrefs}

\begin{thebibliography}{10}

\bibitem{GPML}
C.~E. Rasmussen and C.~K.~I. Williams, {\em Gaussian Processes for Machine
  Learning}.
\newblock MIT Press, 2006.

\bibitem{Williams1998GPC}
C.~K.~I. Williams and D.~Barber, ``Bayesian classification with gaussian
  processes,'' {\em IEEE Transactions on Pattern Analysis and Machine
  Intelligence}, vol.~20, pp.~1342--1351, Dec 1998.

\bibitem{Girolami05VBM}
M.~Girolami and S.~Rogers, ``Variational bayesian multinomial probit regression
  with gaussian process priors,'' {\em Neural Computation}, vol.~18, p.~2006,
  2005.

\bibitem{Kim2006emep}
H.-C. Kim and Z.~Ghahramani, ``{B}ayesian {G}aussian process classification
  with the {EM}-{EP} algorithm,'' {\em IEEE Transactions on Pattern Analysis
  and Machine Intelligence}, vol.~28, pp.~1948--1959, 2006.

\bibitem{Chu2005GPOR}
W.~Chu and Z.~Ghahramani, ``Gaussian processes for ordinal regression,'' {\em
  Journal of Machine Learning and Research}, pp.~1--48, 2005.

\bibitem{Teh05SPLFM}
Y.~W. Teh, M.~Seeger, and M.~I. Jordan, ``Semiparametric latent factor
  models,'' in {\em Artificial Intelligence and Statistics (AISTATS)}, 2005.

\bibitem{Chai2009MGP}
K.~M. Chai, ``Generalization errors and learning curves for regression with
  multi-task gaussian processes,'' in {\em Neural Information Processing
  Systems} (Y.~Bengio, D.~Schuurmans, J.~Lafferty, C.~K.~I. Williams, and
  A.~Culotta, eds.), pp.~279--287, 2009.

\bibitem{Bonilla2008MGPP}
E.~V. Bonilla, K.~M.~A. Chai, and C.~K.~I. Williams, ``Multi-task gaussian
  process prediction,'' in {\em Neural Information Processing Systems}, 2008.

\bibitem{Boyle2005DGP}
P.~Boyle and M.~Frean, ``Dependent gaussian processes,'' in {\em Neural
  Information Processing Systems}, pp.~217--224, 2005.

\bibitem{Salzmann2010CGP}
M.~Salzmann and R.~Urtasun, ``Implicitly constrained {G}aussian process
  regression for monocular non-rigid pose estimation,'' in {\em Neural
  Information Processing Systems}, 2010.

\bibitem{Chan2011CVPR}
A.~B. Chan and D.~Dong, ``Generalized gaussian process models,'' in {\em IEEE
  Conf. Computer Vision and Pattern Recognition}, 2011.

\bibitem{Minka2001}
T.~Minka, {\em A family of algorithms for approximate Bayesian inference}.
\newblock PhD thesis, Massachusetts Institute of Technology, 2001.

\bibitem{Kapoor2010objects}
A.~Kapoor, K.~Grauman, R.~Urtasun, and T.~Darrell, ``Gaussian processes for
  object categorization,'' {\em International Journal of Computer Vision},
  vol.~88, pp.~169--199, 2010.

\bibitem{Feng2004mbrm}
S.~Feng, R.~Manmatha, and V.~Lavrenko, ``Multiple bernoulli relevance models
  for image and video annotation,'' in {\em IEEE Conf. Computer Vision and
  Pattern Recognition}, 2004.

\end{thebibliography}

\end{document}